# From Hand-Perspective Visual Information to Grasp Type Probabilities: Deep Learning via Ranking Labels

Mo Han
Northeastern University
Boston, MA, USA
han@ece.neu.edu

Sezen Yağmur Günay
Northeastern University
Boston, MA, USA
gunay@ece.neu.edu

İlkay Yıldız
Northeastern University
Boston, MA, USA
yildizi@ece.neu.edu

Paolo Bonato
Spaulding Rehabilitation Hospital
Boston, MA, USA
PBONATO@mgh.harvard.edu

Cagdas D. Onal
Worcester Polytechnic Institute
Worcester, MA, USA
cdonal@wpi.edu

Taşkın Padır
Northeastern University
Boston, MA, USA
t.padir@northeastern.edu

Gunar Schirner
Northeastern University
Boston, MA, USA
G.Schirner@northeastern.edu

Deniz Erdoğmuş
Northeastern University
Boston, MA, USA
erdogmus@ece.neu.edu

## ABSTRACT

Limb deficiency severely affects the daily lives of amputees and drives efforts to provide functional robotic prosthetic hands to compensate this deprivation. Convolutional neural network-based computer vision control of the prosthetic hand has received increased attention as a method to replace or complement physiological signals due to its reliability by training visual information to predict the hand gesture. Mounting a camera into the palm of a prosthetic hand is proved to be a promising approach to collect visual data. However, the grasp type labelled from the eye and hand perspective may differ as object shapes are not always symmetric. Thus, to represent this difference in a realistic way, we employed a dataset containing synchronous images from eye- and hand- view, where the hand-perspective images are used for training while the eye-view images are only for manual labelling. Electromyogram (EMG) activity and movement kinematics data from the upper arm are also collected for multi-modal information fusion in future work. Moreover, in order to include human-in-the-loop control and combine the computer vision with physiological signal inputs, instead of making absolute positive or negative predictions, we build a novel probabilistic classifier according to the Plackett-Luce model. To predict the probability distribution over grasps, we exploit the statistical model over label rankings to solve the permutation domain problems via a maximum likelihood estimation, utilizing the manually ranked lists of grasps as a new form of label. We indicate that the proposed model is applicable to the most popular and productive convolutional neural network frameworks.

## CCS CONCEPTS

• **Computing methodologies** → **Machine learning approaches**; **Neural networks**; • **Human-centered computing** → *Interaction devices*.

## KEYWORDS

probability estimation, label ranking, grasp dataset, grasp classification, multi-class classification, prosthetic hand

## 1 INTRODUCTION

To enhance the life quality of an individual who has experienced an upper limb loss, many research groups focus on the development of robotic prosthetic hands to provide a dexterous experience [11, 12]. The newest myoelectrical activity-based designs of the prosthetic hand demonstrate promising results for patients with hand and wrist amputation, but the quality of the muscle activity signals decreases dramatically as amputation severity increases. Recent evidence indicates that deficient electromyogram (EMG) activity could be compensated by electroencephalogram (EEG) signals [1, 14, 19, 22], but because of the lower signal-to-noise ratio of EEG data, the results are still not sufficient for real-world problems. Additionally, frequent calibration of the system is required to account for sensor sensitivity to external factors, such as electrode locations and skin variabilities. Since each patient has a different level of amputation, transferring the learning between subjects is also challenging. The usage of computer vision to drive a robotic controller avoids the aforementioned issues associated with the physiological signals.

With the development of convolutional neural network (CNN) architectures [8, 16, 18, 25, 26, 28, 29], recent studies on computer vision-based prosthetic hand implementation [7, 10–12, 27] mainly focus on grasp detection problems of training CNNs to recognize objects into a grasp type. These studies mostly use an eye-level camera worn by the user (usually located on the user's head), which reduces the convenience and aesthetics while increasing the cost due to the extra visual device. Furthermore, the movement of the human hand and head may not be synchronous, which could cause the inaccurate capture of objects; also the real-time communication between the prosthetic hand and the remote head-camera is another challenge.

In [7] a camera is embedded in the palm of a prosthetic hand as an alternative, where a CNN was trained on *eye-level* images with labels derived manually from *eye-perspective*. However, the identified grasp types for the same object from eye and hand perspectives are usually different for novel objects with irregular shapes. Labellers could directly decide the grasp type to hold an object by observing the hand location and approach direction, but a robotic hand with a palm camera could only view objects from the *hand-perspective*. Therefore, a prosthetic hand design which includes a camera mounted in the palm should be provided a method to train a deep neural network using data from *hand-perspective* as a surrogate for this missing human-like intuition. We collected a customized dataset and trained CNN models using images from the *hand-view*, and also asked labellers to *order the grasp types according to their relevance* for each object from different approaching orientations of *hand perspectives*, instead of annotating a specific object with a fixed *gesture*, i.e. grasp type. To the best of our knowledge, the existing studies use only *eye-view data* for *both* collecting labels and training the CNN, and none of them record data from different object orientations or provide the corresponding EMG and other motion data. All of these aspects are covered in this dataset.

Visual information is easy to obtain and relatively stable to users and environmental variations, and thus leads to more intuitive decisions as well as robust results. However, building a prosthetic hand with only computer vision techniques and excluding the human effect may not be the best option. Physiological signals are still required to detect the human response and intention. For instance, moving towards a closer object may indicate the intention to push it away in order to grasp a further one. Therefore, a more reliable solution to a prosthetic hand design should leverage the strengths of both visual information and biomedical signals.

To facilitate multi-modal fusion of computer vision with physiological signal inputs, a probability-based classifier is preferred over an absolute prediction. A general idea to estimate multi-class distributions is to transform ranking scores of labels into accurate probability estimates while minimizing the overall classification error [2, 5, 6, 15, 17, 21, 31, 32]. In [2, 5, 15, 31, 32], these scores are generated from the distances to the decision hyper-planes of different classes learned by support vector machines or k-nearest neighbors algorithm. Other score metrics are also applied to this problem in [6, 17, 21], such as Kendall distance, surrogate loss, Hamming loss, etc. However, mostly these methods are only for binary classification, or more prevalently, they divide multi-label learning into multiple independent binary classification problems. Thus, most of these approaches suffer from imbalanced data distributions while building binary classifiers to distinguish the correct class. With the increase in the number of classes, this problem gets even worse, and simultaneously the computation also shows a squared growth. Therefore, in this paper, we apply an alternative approach which undermines these disadvantages, based on the Plackett-Luce (PL) Model [3, 4, 13, 20, 23]. The general idea is to estimate the probability models on rankings to optimize permutation domain problems. This problem could be solved through a maximum likelihood estimation, by bridging the distribution estimation to label ranking. Therefore, to address the label ranking, we manually list the permutation of grasps as a novel form of label for the collected image data: instead of giving purely positive or negative labels, the labellers were asked to order grasp types from the most relevant to the most irrelevant ones. Compared to the typical classification problems, our task is not detecting one certain label but deciding the distribution over grasps and also the best grasp order using the labels collected from different users who may have alternative preferences. Thus, considering the difficulty of the problem, the results are promising. Additionally, the study shows that the proposed model is applicable to the most popular and productive CNN frameworks.

## 2 NOVELTY

### 2.1 Eye-View Images for Labeling and Hand-View Images for Training

In our dataset, the visual data (both image and video) of various objects from both hand and eye views were collected simultaneously, since the footage of the palm camera while approaching the object does not match with the eye aspect due to the nature of the problem. Therefore, to imitate this human-like pattern for a palm camera, instead of training models with eye-view images, we used the hand-perspective data of the dataset as training images, and manually annotated these hand-view pictures by showing eye-level images to labellers. It is worth to stress that the eye-level images were only for manual labelling because the labels are only decided by the human eye-view of observing the hand location with respect to the object, while a CNN prepared for a robotic hand could be trained with hand-view images because of the visual pattern of the hand. Furthermore, the visual data were collected from different orientations of the objects to learn the complex shape information of each target, rather than assigning one fixed label to each object. Electromyogram activity and dynamic data from the upper arm were also recorded for further studies.

### 2.2 *Ranking Label*

As introduced in the following sections, the distribution estimate of all possible grasp types could be implemented by introducing probability models on rankings to optimize permutation domain problems. To do so, during the data collection, instead of picking the most relevant hand gesture, the labellers were asked to provide the preferred order of multiple grasp types; the ranked permutation of grasps is called *ranking label*. By exploiting this special form of high-dimensional label, the prediction of a CNN will be a probability distribution over gestures, of which the top one is the most likely gesture of a successful grab. However, existing multi-class datasets

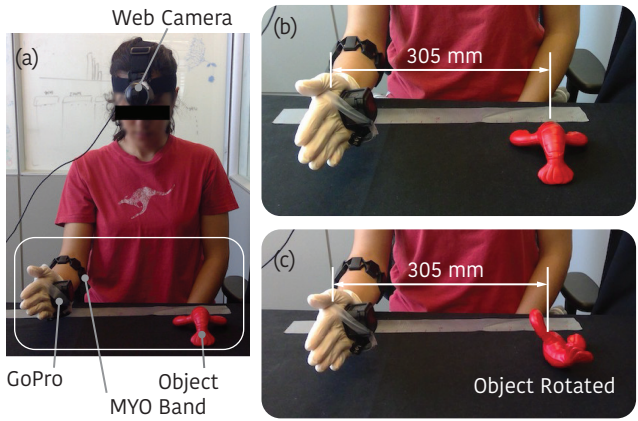

**Figure 1: The set-up of the data collection: (a) an experimenter wore a headband which held a web camera for the human eye-view aspect, whereas, a GoPro camera was attached on experimenter's hand to collect visual information from the hand aspect. The EMG signals and movement kinematics were recorded with a MYO armband. (b) The initial distance between the object and hand was 305 mm, and it was fixed before and right after the approaching. (c) The object was horizontally and vertically rotated to capture all possible orientations.**

usually contain labels which are definitely positive or absent but still considered as positive.

## 2.3 Probability Estimation Via Label Ranking

Different from the common classification tasks to decide one classification out of several classes, in order to facilitate the combination of computer vision with the EEG and EMG systems in the future, and also to reflect the diversity of user preferences on different grasps, we build a novel model for estimating the entire statistical distribution over multiple classes and then ranked them, by making use of the *ranking label* introduced above. The multi-class ranking and probability distributions could be bridged by the Plackett-Luce (PL) model, which introduces a probability model on rankings to optimize permutation domain problems. Then, by introducing a maximum likelihood estimation problem, the prediction of the probability distribution of gestures could be solved by label ranking. By rewriting the estimated probability distribution of the multiple classes with the order of the most to the least likely, the estimated ranked list of labels could also be obtained. This model is applicable to other types of implementations for distribution estimates and ranking proposes, and it is proved in the experiments that the most common and productive CNN structures such as GoogLeNet, InceptionV3 and ResNet50 are able to be utilized as the base network.

## 3 DATA COLLECTION

In this section, we explain the dataset collected by the authors, which is for the probability estimation and ranking of multiple grasps, for the prosthetic hand with a camera mounted in the palm.

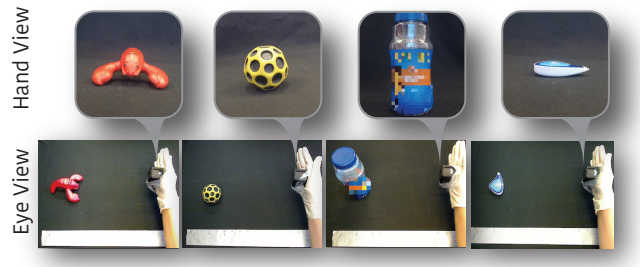

**Figure 2: Eye-view images were captured by a webcam (Logitech Webcam C600, 1600×1200 resolution) for labelling and corresponding hand-view images were captured by GoPro (GoPro Hero Session, resolution of 3648×2736 pixels) for training.**

### 3.1 Set-up and Collection

The visual data (both image and video) of various objects from both hand and eye views were collected simultaneously. As demonstrated before, considering the natural behavior pattern of human, the eye-view photos were only provided to labellers for manual annotation instead of training CNNs; while these labels and corresponding hand-view images were the ones fed into the CNN for training. Furthermore, to adapt to the irregularity of the object shape, the visual data were collected from different orientations for each object. To enlarge the dataset, augmentation (introduced in Section 3.3) were applied to the hand-view photos. Additionally, the dataset includes videos from both eye and hand perspectives, EMG signals collected from upper arm, and movement kinematics (acceleration, duration, gyroscope and orientation) collected while approaching the object.

The set-up for data collection is shown in Fig. 1. During data collection, an experimenter wore a headband which held a camera for the human eye-view, while another camera was attached on experimenter's right hand to collect visual information from the hand aspect. First, to collect the static image data, the subject kept the right hand and head stationary, and both cameras captured photos simultaneously (see Fig. 2). Then, with an audio cue, the experimenter started to approach the object while videos, EMG, and movement kinematic acquisition devices started to collect all trials. The record was stopped when the experimenter finished grasping the object and returned to the initial state, which was 305 mm (12 inches) far from the object. The same procedure was repeated a number of times while the object was rotated vertically and horizontally to capture all possible orientations. In order to simplify the vision problem, all images were taken in front of a black curtain.

The current dataset consists of 4466 images and labels for training, which were generated from 413 hand-perspective pictures of 102 ordinary objects, including office and daily supplies, utensils, and complex-shaped objects like stuffed animals. In addition, for the potential applications in the future, the dataset includes videos from both eye and hand perspectives, EMG signals and corresponding movement kinematics, where the videos could be utilized to understand the distance between the hand and object to decide the time when the hand needs to start pre-shaping to meet object demands, and EMG and kinematics signals could be used for a hybrid multi-modal controller.

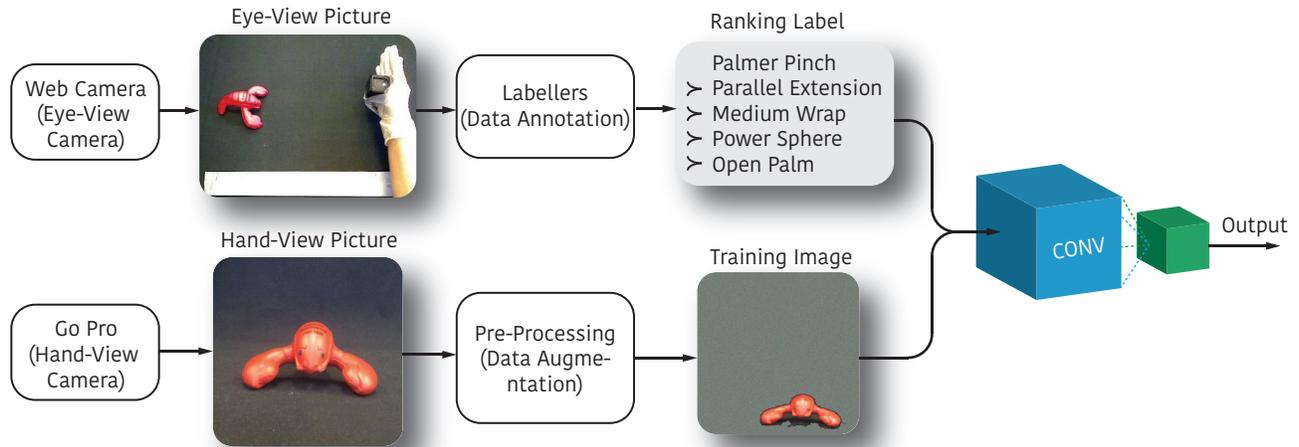

Figure 3: The usage of images from eye and hand perspectives: First, the eye- and hand- perspective photos were taken simultaneously. Then, the eye-perspective images were shown to labellers for the manual annotation to obtain the ordered list of 5 gestures. At the same time, the pictures from hand-view were pre-processed, which were then matched with the corresponding labels, by which a CNN model was trained.

Even though human grasp types could be grouped into 33 classes in [9], some of them are too complex for a robotic implement. As proved in [9], some of the grasp tasks are very similar with the others and most of them could be handled successfully by several ones of all the 33 gestures. Moreover, it has been shown in [22] that, EEG signals are only capable of classifying 5 specific grasp types. Thus, for the future fusion with the EEG control, the label set is limited to 5 gestures: Open Palm, Medium Wrap, Power Sphere, Parallel Extension and Palmar Pinch, as shown in Fig. 4.

### 3.2 Data Annotation: *Ranking Label*

To estimate the probabilistic distribution and ranking of the gestures, each image was annotated by a *ranking label* – an ordered permutation of 5 grasp types rather than a single grasp type. In order to cover the human decision variability and combat the biases of different subjects, 11 individuals participated in the label collection (which was conducted separately from the data collection), and they were randomly shown eye-view images through an interface. Then they were asked to order the 5 grasp types for each eye-view image, which could be denoted as a bijection of $(g_1, \cdots, g_5)$ onto $(g'_1 > \cdots > g'_5)$, where $g_i$ and $g'_i$ are grasp classes. For example, a possible *ranking label* could be 'Palmar Pinch > Parallel Extension > Medium Wrap > Power Sphere > Open Palm'. Each sorted ranking is a 5-dimension label, which was then paired with its corresponding training (hand-view) image to train CNNs after pre-processing.

### 3.3 Data Augmentation and Pre-processing

Similar to any deep neural network model, CNNs require relatively large datasets for training in order to avoid overfitting. To satisfy this requirement, each image was matched with multiple labels obtained from different individuals. Additionally, pre-processing methods were applied to the raw hand-perspective images to get rid of the effect from the image background. First, the objects were segmented out of the original image and randomly blurred. Then, Gaussian noise with random variance was added in the background to make the network more robust to the various real-world backgrounds. Finally, the blurred object was randomly located in the noisy background to form the final training image. The bounding box information of the object location was also recorded as a part of the true label, which has a potential value for indicating the ground truth to any object detection study. The complete process of image collection and usage is shown in Fig. 3.

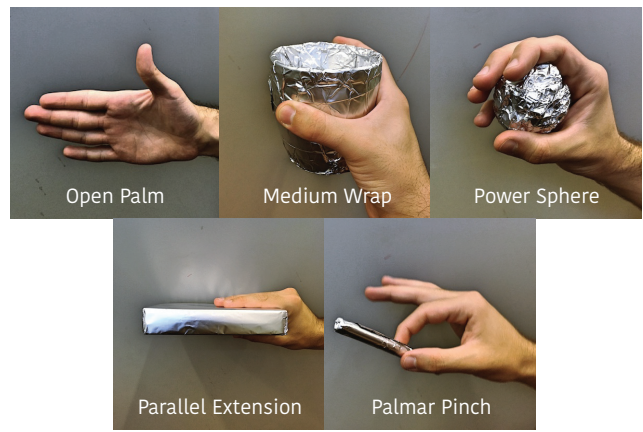

Figure 4: The selected grasp types.

## 4 METHODOLOGY

The proposed model for estimating probability distributions and multi-class ranking is applicable to the most common CNN structures. The study utilizes several popular deep CNN structures as the base network to extract features from images, and builds a model to solve the distribution estimation problem by exploiting the *ranking label* and visual information through a customized loss function on the basis of the Plackett-Luce Model for multi-class label ranking. In this section, the aforementioned implementation details are explained.

### 4.1 Learning Models of CNN

In this work, we apply efficient image-classification frameworks including GoogLeNet [28], InceptionV3 [29] and ResNet50 [16] as the examples of a base network structure for comparison. The adopted CNNs here have relatively small model size but deep network structure, which could obtain high accuracy with fast training speed and low overfitting [29]. GoogleNet, which is also called InceptionV1, has 22 layers, 9 inception modules and 7 million parameters. It requires fewer parameters than its previous structures while providing higher accuracy. The inception layer could cover a bigger area, but also keep a fine resolution for detailed information on the images. The InceptionV3 is a 42-layer deep learning network of 23 million parameters, with additional batch normalization and factorization ideas introduced to reduce overfitting. The ResNet is the abbreviation of Residual Network, which learns the residual of skipping connections and thus enables the development of much deeper networks. ResNet50 has 50 layers and 25 million parameters.

### 4.2 Maximum Likelihood Estimation for Ranking Label

The considered true label here to train CNNs is the *ranking label*, which is a 5-dimensional ranked list of grasps. Therefore, we define a proper loss function for the CNN to extract information from this high-dimensional label. The aim of the loss function is to maximize the probability of a specific 5-dimensional permutation among all possible permutations, rather than maximizing the probabilities of one or more classes, i.e. grasps, in the list. The probability of a ranking is calculated by the Plackett-Luce model, which links the *ranking label* to a probability distribution and thus leads to our loss function for the problem.

*4.2.1 The Plackett-Luce Model.* The Plackett-Luce (PL) model establishes a natural bridge between label ranking and probability distributions [20, 23]. In this model, $n$ elements of interest have a corresponding probability distribution $\boldsymbol{\omega} = (\omega_1, \omega_2, \cdots, \omega_n)$ to be estimated, satisfying $\sum_{i=1}^{n} \omega_i = 1$. The distribution $\boldsymbol{\omega}$ is related to the probability of a possible ranking via:

$$P(\pi^k|\boldsymbol{\omega}) = \prod_{i=1}^{n-1} \frac{\omega_{\pi^k \langle i \rangle}}{\sum_{j=i}^{n} \omega_{\pi^k \langle j \rangle}}, \tag{1}$$

where $\boldsymbol{\pi} = \{\pi^1, \pi^2, \cdots, \pi^N\}$ is a set of independent rankings of $n$ elements. Here, $N$ is the total number of the collected *ranking labels*.

Each permutation $\pi^k$ for $k \in \{1, 2, \cdots, N\}$ represents a possible ranking for the $n$ elements, where $\pi^k \langle i \rangle$ is the top $i^{th}$ element among all items.

*4.2.2 Maximum Likelihood Estimation Based on PL Model.* The maximum likelihood estimate of the probability distribution $\boldsymbol{\omega}$ is given by those parameters that maximize the likelihood in Eq.(1) or, equivalently, the following log-likelihood function:

$$\ell(\boldsymbol{\omega}) = \sum_{i=1}^{n-1} \left( \ln \omega_{\pi^k \langle i \rangle} - \sum_{j=i}^{n} \ln \omega_{\pi^k \langle j \rangle} \right). \tag{2}$$

In our problem, there are $n = 5$ grasp types to be ordered. Accordingly, the $k^{th}$ *ranking label* can be demonstrated as

$$\begin{aligned} \pi^k &= \left( \pi^k \langle 1 \rangle, \pi^k \langle 2 \rangle, \cdots, \pi^k \langle 5 \rangle \right) \\ &= \left( \pi^k \langle 1 \rangle \succ \pi^k \langle 2 \rangle \succ \cdots \succ \pi^k \langle 5 \rangle \right), \end{aligned} \tag{3}$$

where $\pi^k \langle i \rangle$ can be any gesture shown in Fig. 4. Therefore, the estimated probability distribution $\hat{\boldsymbol{\omega}}$ can be expressed as:

$$\begin{aligned} \hat{\boldsymbol{\omega}} &= \arg \max \ell(\boldsymbol{\omega}) \\ &= \arg \min loss, \end{aligned} \tag{4}$$

where $loss$ is the loss function defined as

$$loss = -\ell(\boldsymbol{\omega}), \tag{5}$$

and $\boldsymbol{\omega} = (\omega_1, \omega_2, \cdots, \omega_5)$ is the true distribution for 5 grasps. Denoting the estimated distribution $\hat{\boldsymbol{\omega}} = (\hat{\omega}_1, \hat{\omega}_2, \cdots, \hat{\omega}_5)$ as:

$$\hat{\boldsymbol{\omega}} = (\hat{\omega}_{\hat{\pi}^k \langle 1 \rangle} \succ \hat{\omega}_{\hat{\pi}^k \langle 2 \rangle} \succ \cdots \succ \hat{\omega}_{\hat{\pi}^k \langle 5 \rangle}), \tag{6}$$

for the $k$th input, we obtain the estimated ranking of all labels as:

$$\begin{aligned} \hat{\pi}^k &= \left( \hat{\pi}^k \langle 1 \rangle, \hat{\pi}^k \langle 2 \rangle, \cdots, \hat{\pi}^k \langle 5 \rangle \right) \\ &= \left( \hat{\pi}^k \langle 1 \rangle \succ \hat{\pi}^k \langle 2 \rangle \succ \cdots \succ \hat{\pi}^k \langle 5 \rangle \right). \end{aligned} \tag{7}$$

Thus, by estimating the probability distribution $\hat{\boldsymbol{\omega}}$, we simultaneously solve the ranking problem for $n = 5$ classes.

## 5 EXPERIMENTS AND EVALUATIONS

### 5.1 Experiment Setup

*5.1.1 Dataset.* We evaluate the proposed model on our custom dataset introduced in Section 3. In total, 4466 pairs of hand-view image and its corresponding *ranking label* are included in the experiment, which are generated from 413 hand-perspective pictures and 11 labellers. The model is trained by 80% portion of those unseen data-pairs from objects randomly selected, whereas, tested on the remaining 20% images and labels from other objects. All input images are resized into the dimension of $224 \times 224$. In the *ranking label*, 5 grasp types are contained as shown in Fig. 4: Open Palm, Medium Wrap, Power Sphere, Parallel Extension and Palmar Pinch.

*5.1.2 Training Setup.* Here we use GoogLeNet, InceptionV3 and ResNet50 without their last fully connected layers as our base network respectively, and leverage transfer learning to initialize these structures with their pre-trained weights on ImageNet [8]. On top of this, we build fully connected layers whose output is a 5-dimensional vector with sigmoid activations. We train these architectures with stochastic gradient descent, where the learning rate ranges from $10^{-5}$ to $10^{-3}$. To comparatively test the performance of coping with overfitting, for the GoogLeNet structure, we add L2 regularizers with a regularization parameter $\lambda$ in each convolutional layer, where $\lambda$ ranges from 0.0002 to 0.02, among which 0.002 is the default value given by official GoogLeNet model; we choose two more values around the default parameter to test the model performance. We define our own loss function as (5) to estimate the multi-class probability distribution by learning the *ranking label*.

*5.1.3 Accuracy to Evaluate Ranking.* The original output of the trained CNNs is an estimated probability distribution of 5 dimensions: $\hat{\omega} = (\hat{\omega}_1, \hat{\omega}_2, \cdots, \hat{\omega}_5)$. Nevertheless, the benchmark to evaluate the prediction is the collected *ranking label* with the form of (3). Therefore, we compare the estimated permutation (7) with the true *ranking label* to evaluate our model.

There are two common ways to quantify the similarity between two sorted lists: 1) rank correlation-based method, 2) set-based measure. The first method essentially measures the probability of two items being in the same order in the two ranked lists, where the exact position of an item has no effect on the final similarity score, such as Kendall's tau coefficient [24]. However, in practice, we are more concerned with the grasps of higher relevance, which are the top rankings among all grasps. Therefore, we utilize the second approach ([30]) to compare two permutations. Generally, this method finds the average fraction of the content overlapping of subsets with different depths. This average overlap score, which is defined as the accuracy for our problem, is calculated as follows:

$$acc = \frac{\sum_{i=1}^{n} f_i}{n} \quad (8)$$

where $n$ is the total number of labels in the list, and $f_i$ is the fraction of the content overlapping over the subsets of the top $i$ rankings from the two ranked list to be compared, i.e. $f_i = \frac{i'}{i}$ where $i' \leq i$ is the number of overlapping items of the top $i$ rankings between two ranked lists. Since observing a common item at higher rank contributes to all the lower ranked subsets, this approach is naturally top-weighted, i.e. it assigns higher relevance to items at the higher ranks. For instance, we compare the lists of A: $\{a, b, c\}$ and B: $\{a, c, b\}$ with $n = 3$, and their subsets with depth $i = 1$ are $\{a\}$ and $\{a\}$ respectively, where the number of overlapping item $i' = 1$ and $f_i = \frac{1}{1} = 1$. When $i = 2$ the subsets of A and B are $\{a, b\}$ and $\{a, c\}$, where $i' = 1$ and $f_i = \frac{1}{2}$. Similarly, for depth $i = 3$, the subsets are A and B themselves, with all the $i' = 3$ items overlapping and $f_i = \frac{3}{3} = 1$. The final accuracy can be calculated as $acc = (1 + \frac{1}{2} + 1)/3 = 83.33\%$.

## 5.2 Results and Discussion

*5.2.1 Results.* We train the proposed model on our dataset with GoogLeNet, InceptionV3 and ResNet50 as base networks, and evaluate the trained CNNs through the accuracy defined as (8). Table 1 presents the prediction performances of models with structures of GoogLeNet, InceptionV3 and ResNet50, where the accuracy of GoogLeNet is averaged over different $\lambda$. Table 2 shows the prediction performances of our model based on GoogLeNet structure with respect to different regularization parameters. For each model, we determine the best learning rate (L.R.) by comparing their accuracy. Specifically for the GoogLeNet framework where L2 regularizers are embedded in each convolutional layer, we determine the optimal regularization parameter $\lambda$.

Overall, the accuracy of 15 trained models is around 86%. The best model among all trained CNNs is the structure of GoogLeNet when $\lambda = 0.002$ and $L.R. = 10^{-3}$, where the accuracy is 87.1836%. It implies that given an object image, in average over 87% portion of its corresponding ranked list of grasps could be predicted accordingly, where the grasps with higher ranks contribute more to the accuracy than the lower ones. According to Table 1, compared with InceptionV3 and ResNet50, GoogLeNet performs relatively better even with much fewer parameters (as introduced in Section 4.1), partly because of its embedded L2 regularizers, which reduce the overfitting to training data. The training speed of GoogLeNet is also 2 times higher than the other CNNs due to its smaller parameter size. Moreover, InceptionV3 demonstrates slightly better performance than ResNet50. The reason for this difference could be that InceptionV3 could deal with detailed information better, and the grasping targets are always small in size. For all 3 CNN frameworks, the learning rate of $L.R. = 10^{-3}$ leads to their optimum performances. Specifically, on the basis of Table 2, it is shown that the optimal parameters for L2 regularizers and learning rate of GoogLeNet are $\lambda = 0.002$ and $L.R. = 10^{-3}$. The optimum of $\lambda = 0.002$ is same as the default value given by GoogLeNet, and the optimum of $L.R. = 10^{-3}$ is aligned with those of the other two CNNs.

**Table 1: Prediction performances of GoogLeNet, InceptionV3 and ResNet50: L.R. indicates learning rate. The accuracy of GoogLeNet is averaged over different regularization parameters $\lambda$. GoogLeNet leads to better performances than InceptionV3 and ResNet50; learning rate of $L.R. = 10^{-3}$ leads to the optimums of all 3 frameworks. InceptionV3 obtains better performance than ResNet50.**

| L.R. | Accuracy | | |
|---|---|---|---|
| | GoogLeNet | InceptionV3 | ResNet50 |
| $10^{-5}$ | 86.3866% | 86.2048% | 85.7910% |
| $10^{-4}$ | 86.0603% | 86.2027% | 86.1239% |
| $10^{-3}$ | **86.5269%** | **86.4930%** | **86.4142%** |

**Table 2: Prediction performances of GoogLeNet: $\lambda$ indicates the L2 regularization parameter. The optimal parameter is $\lambda = 0.002$ and $L.R. = 10^{-3}$, resulting in the highest accuracy of 87.1836% among all 15 trained models.**

| $\lambda$ | L.R. | Accuracy of GoogLeNet |
|---|---|---|
| | $10^{-5}$ | 86.4080% |
| 0.0002 | $10^{-4}$ | 86.0368% |
| | $10^{-3}$ | 86.1031% |
| | $10^{-5}$ | 86.1944% |
| **0.002** | $10^{-4}$ | 86.0368% |
| | **$10^{-3}$** | **87.1836%** |
| | $10^{-5}$ | 86.5573% |
| 0.02 | $10^{-4}$ | 86.1073% |
| | $10^{-3}$ | 86.2939% |

*5.2.2 Discussion.* Here we note that expecting an accuracy as high as the typical object classification implementations would be misleading due to two apodeictic facts: 1) more difficulties of predicting an ordered list of labels than predicting one individual label; 2) the biases from different labellers and complex object shapes for each image.

With the goal of predicting a distribution and ordering five grasp types, the 100% accuracy of definition (8) means that not only detecting the most possible grasp type for the given image, but also finding the exactly same order of all gestures as the true *ranking label*, i.e. for any two grasps among the 5 gestures, they have to be in the same order as the true label, and there are $C(5, 2) = 10$ pairs of them in total. Thus compared to the typical grasp classification problems, our task is more difficult due to the nature of the problem and thereby results in a more practical accuracy.

On the other hand, the 11 labellers have different grasp preferences for each image. Even if when the same person grasps the same object for multiple times, different gestures may be used unconsciously. In this work, 11 labellers annotated the same image set, which refers that each image could have more than one and up until 11 kinds of *ranking labels* according to the distribution of labellers' preferences, and all of the 11 true labels would be the benchmarks for calculating the average accuracy. Thus when predicting a ranked grasp list for a specific image, an average 100% accuracy requires the prediction to be completely the same with all of the 11 true labels, i.e. the 11 true *ranking labels* have to be exactly the same, which is not feasible. Therefore the accuracy would always include the natural biases from different labellers. However, these biases are meaningful and necessary to reflect the properties of objects with different complexity. For example, considering some simple objects such as a book, it is very obvious to utilize Parallel Extension for grasping, and the orders of the 5 grasps given by different labellers could be very similar, where the biases from labellers are relatively small. In this situation, the predicted possibilities of grasps may reflect a significant decline from the top grasp to the last one. Whereas, for objects with complex shapes such as the hand sanitizer bottle whose different parts may lead to different grasp types, the ranked lists of grasps given by different labellers could be quite different, and thus result in a more uniform predicted distribution over all the grasps. When combined with other evidences (EEG and EMG), the higher uniformity of estimated distributions from visual system, which indicate the higher user biases and object complexity, would contribute less to the final predicted distribution and thus make the entire system dependent more on other signal resources from human interference, leading to more reliable results.

## 6 CONCLUSION

In this paper, we consider the probability estimation problem of multi-class classification based on label ranking for a prosthetic hand design relying on the visual information obtained from different perspectives. Considering the asymmetric shapes of ordinary objects and the visual pattern of robotic hands, we collected a dataset with images of various objects from both hand- and eye-views simultaneously, where the eye-view images were used only for labelling, while the hand-perspective pictures were fed to the CNN after augmentation and preprocessing. The data were collected from different directions, and videos of both perspectives, EMG signals, and movement kinematics of the upper arm were also recorded during the grasp for the future combination of multiple systems. Furthermore, we built a probability-based classifier instead of providing absolute positive or negative predictions for the prosthetic hand design by employing the Plackett-Luce model. By introducing probability models on rankings to optimize permutation domain problems, the PL model links the distribution estimation with the problem of label ranking. By solving a maximum likelihood estimation problem based on this model, a customized loss function was defined for training CNNs to estimate the probability distribution over the ordered labels. The ranked list of grasps was estimated from the predicted distribution by sorting the possibilities of labels. We utilized several efficient CNNs to extract features while implementing and testing the model, and found the best framework and parameters among all trained CNNs by comparing their results. In the future, this work could be further extended to other applications that focus on the distribution estimation and label ranking.

## ACKNOWLEDGMENTS
Our work is supported by NSF (IIS-1149570, CPS-1544895, CPS-1544636, CPS-1544815) and NIH (R01DC009834).